%% file: root.tex
\documentclass[letterpaper, 10 pt, conference]{ieeeconf}  

\IEEEoverridecommandlockouts                              

\usepackage{graphics} 
\usepackage{epsfig} 
\usepackage{mathptmx} 
\usepackage{times} 
\usepackage{amsmath} 
\usepackage{amssymb}  
\usepackage{bm}
\usepackage{balance}
\usepackage{booktabs}
\usepackage[table]{xcolor}
\let\labelindent\relax
\usepackage{enumitem}

\title{\LARGE \bf
SEM: Enhancing Spatial Understanding for Robust  Robot Manipulation
}

\author{
Xuewu Lin, Tianwei Lin, Yun Du, Yiwei Jin, Jitao Li, Hongyu Xie, Lichao Huang, Zhizhong Su*
\thanks{*Horizon, Beijing, China, zhizhong.su@horizon.auto}
}

\begin{document}

\maketitle
\thispagestyle{empty}
\pagestyle{empty}

\begin{abstract}


A key challenge in robot manipulation lies in developing policy models with consistent spatial understanding—the ability to reason about 3D geometry, object relations, and robot state.
Existing mainstream models take 2D images as input, without performing explicit 3D modeling, and thus lack spatial understanding capabilities as well as 3D and embodiment generalization.
To address this, we propose SEM (Spatial Enhanced Manipulation), a diffusion-based policy framework that constructs a unified spatial representation by projecting multi-view image features and joint-centric robot states into a unified 3D space.
This spatially aligned representation provides a consistent feature space for the diffusion policy to condition on during action generation. 
Extensive experiments demonstrate that SEM significantly improves spatial understanding, leading to robust and generalizable manipulation across diverse tasks that outperform existing baselines.

\end{abstract}

\section{INTRODUCTION}

Robust robot manipulation in diverse real-world environments hinges on a core capability: spatial understanding—the ability to perceive 3D geometry, object relations, and the robot's own embodiment states in a consistent way.
While recent diffusion-based (or flow-based) policy models have demonstrated strong generative capacity for action modeling, their effectiveness remains limited by perceptual bottlenecks:
(1) vision backbones are typically trained for 2D image understanding (visual grounding and visual question answering) and lack explicit 3D reasoning ability,
(2) visual features and robot state features are processed independently without spatial alignment, and
(3) the robot’s embodiment structure is not effectively modeled.

\begin{figure}
    \centering
    \includegraphics[width=1.0\linewidth]{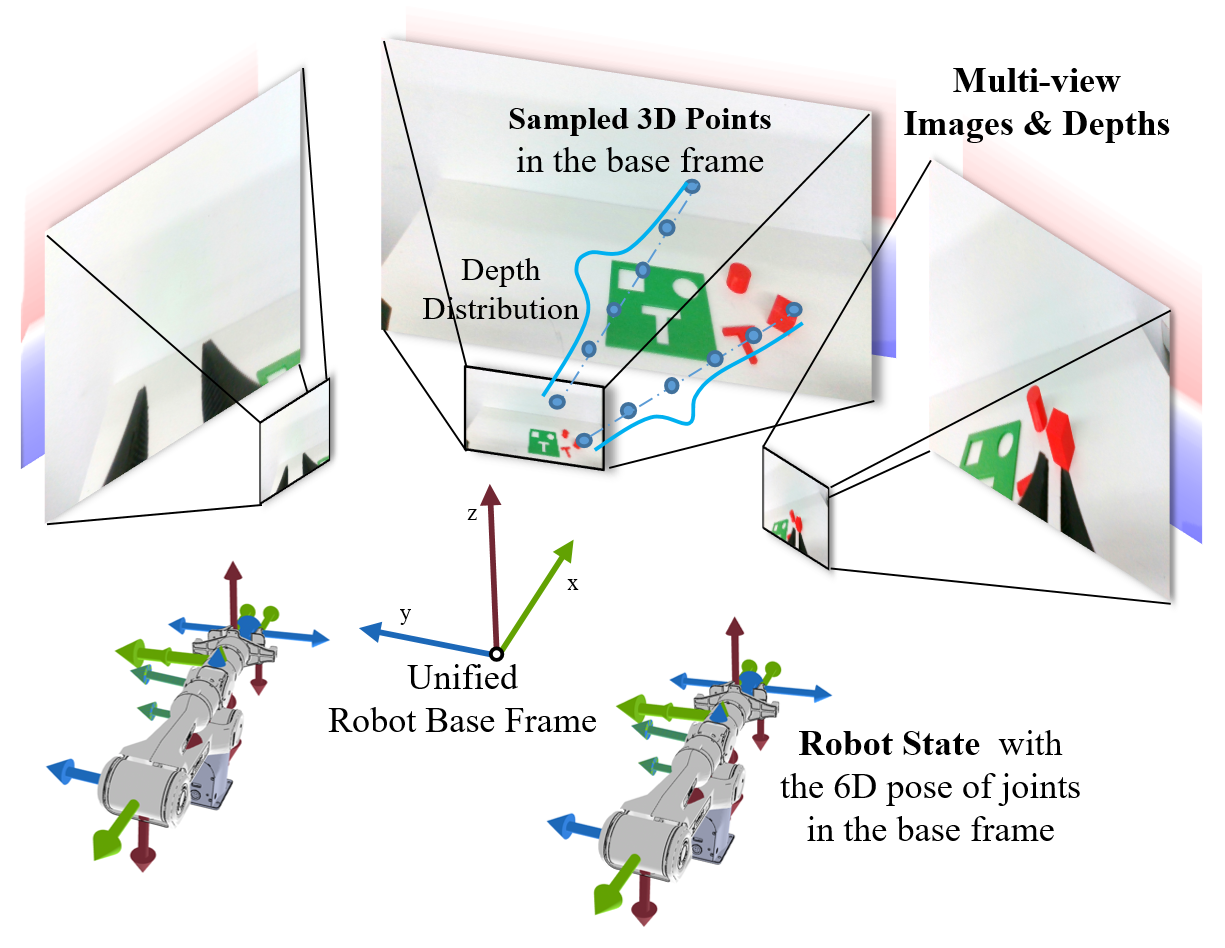}
    \caption{
    Overview of SEM. SEM constructs a unified spatial representation by projecting multi-view image features and joint-level robot states into the robot base frame.
    }
    \label{fig:overview}
\end{figure}

To address these challenges, we propose SEM (Spatial Enhanced Manipulation), a policy framework that constructs a unified spatial representation as a consistent representation space for action generation (Fig. \ref{fig:overview}).
%
First, we introduce a 3D spatial enhancer that injects explicit 3D information into 2D visual features. This module encodes camera intrinsic and extrinsic parameters into the visual representation via 3D absolute position embeddings.
%
Second, when encoding the robot state, we augment the joint angles with the 6D poses, obtained via forward kinematics, and embed these poses as part of the robot state features. Both the 3D absolute position of visual features and the joint poses are expressed in a unified 3D coordinate frame, such as the robot base frame.
%
Finally, to better capture the embodiment structure and support robots with arbitrary numbers of joints, we design a joint-centric robot state encoder and action decoder, which enable feature propagation among joints through graph attention.


Robot embodiments and sensor configurations often differ significantly across datasets. When training policies on such heterogeneous data, severe ambiguities can arise: the same joint index may correspond to distinct physical motions on different robots, and varying camera setups can yield observations from inconsistent viewpoints.
These ambiguities are a major obstacle to embodiment and sensor generalization in existing models.
These ambiguities are the key factor limiting the embodiment and sensor generalization capabilities of existing models. SEM mitigates such inconsistencies by explicitly modeling both the camera and the robot embodiment, thereby achieving 3D and embodiment generalization and obtaining improved performance.

%

We extensively evaluate SEM in both simulation and real-robot settings.
In simulation, we adopt RoboTwin 2.0 benchmark~\cite{robotwin2}  and evaluate on 16 representative tasks.
Despite having only around 40M trainable parameters, SEM significantly outperforms existing approaches.
We further conduct targeted experiments on 3D spatial generalization and embodiment generalization, confirming the benefits of our unified spatial representation.
In real-world evaluations, SEM consistently outperforms RDT~\cite{rdt} across all seven tasks.
Our contributions are summarized as follows:
\begin{itemize}[leftmargin=8pt]
    \item {\bf Novel Framework.}
    We propose SEM, a new VLA framework, from the perspective of enhancing spatial understanding.
    \item {\bf Superior Performance.} We evaluate SEM thoroughly in both simulation and real-world robot settings, demonstrating that SEM outperforms existing baselines.
    \item {\bf Comprehensive Analysis.} We verify the effectiveness of each module in SEM, as well as its improvement in 3D spatial and embodiment generalization.
\end{itemize}

\section{Related Works}
\label{sec:related_works}

End-to-end robot manipulation methods have progressed rapidly, with improvements being made from various perspectives.
%
Early works \cite{1988,bco,deepimitationlearning,zeng2020tossingbot} use imitation learning to acquire manipulation skills from human demonstrations. Subsequent studies refine action modeling to improve stability and performance: IBC \cite{IBC} introduces implicit behavioral cloning, BeT \cite{BeT} models multi-modal actions with a Gaussian mixture, and Diffusion Policy (DP) \cite{dp} formulates trajectory generation as a conditional diffusion process, enabling robust learning in high-dimensional action spaces.

Moreover, many works have contributed to the improvement of algorithmic pipelines and network architectures.
PerAct \cite{peract} encodes voxelized 3D features with a Perceiver transformer to predict end-effector (EE) keyframes. 
Act3D \cite{gervet2023act3d} introduces coarse-to-fine 3D point sampling for precise action prediction. 
3D Diffuser Actor \cite{3ddiffuseract} combines Act3D with diffusion modeling for further gains. 
RVT \cite{goyal2023rvt} re-renders observations into three orthogonal planes to produce heatmaps of EE keyframes, while RVT-2 \cite{goyal2024rvt-2} adds a fine branch to boost accuracy. 
KStarDiffuser \cite{kstar} uses graph convolution over joint trajectories with an auxiliary joint position loss for bimanual tasks. However, all these approaches predict EE poses, which limits their ability to handle complex long-horizon or contact-rich tasks such as cloth manipulation. DP3 \cite{dp3} addresses this by directly predicting joint positions via diffusion, enabling finer control.

 \begin{figure*}[tb]
     \centering
     \includegraphics[width=0.92\linewidth]{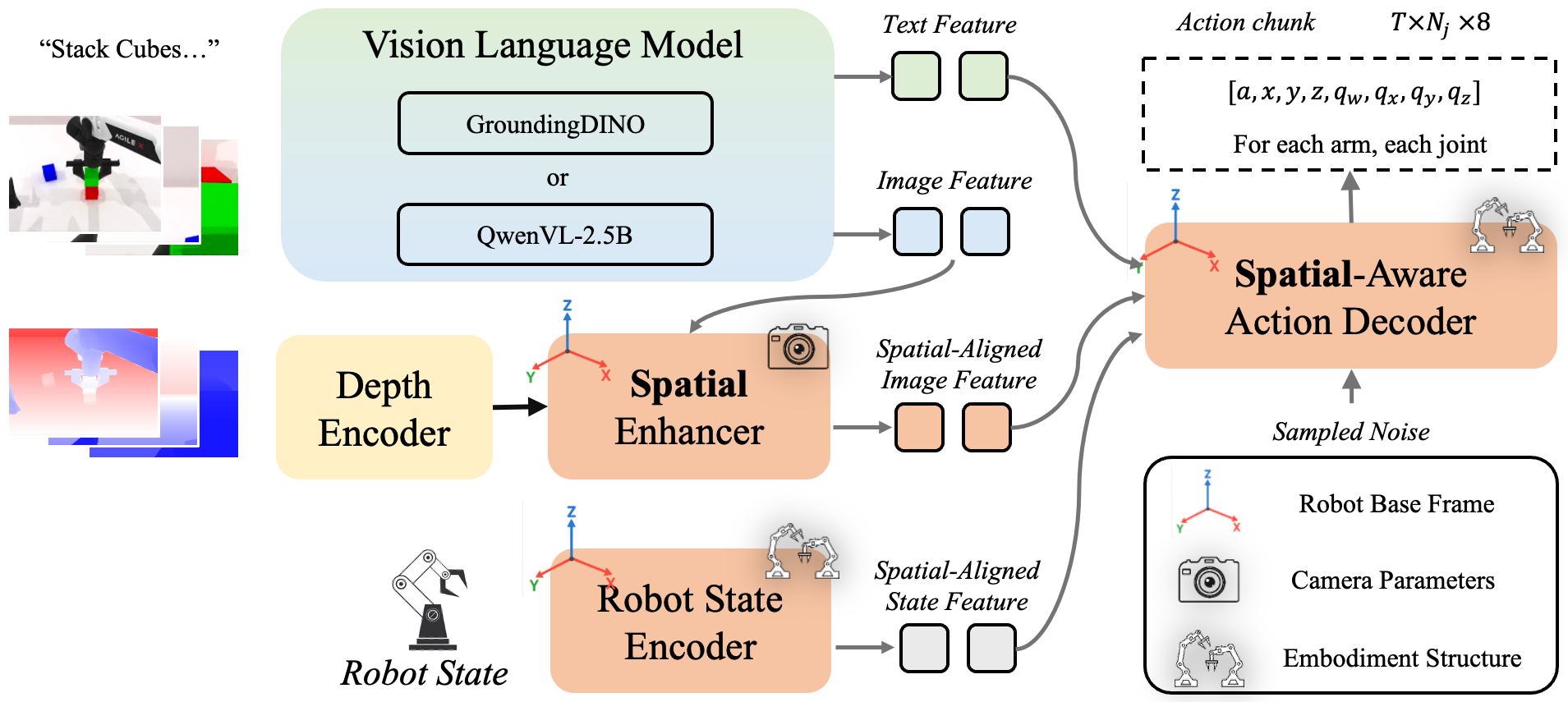}
     \caption{
     The overall architecture of SEM. SEM takes multi-view images, depth, robot state, and instructions as inputs, projects them into a shared base-frame 3D space, and generates future joint trajectories via a diffusion-based policy. 
     In this way, SEM learns a spatially consistent representation that improves performance and generalization.
     }
     \label{fig:overall_struture}
 \end{figure*}

Recently, breakthroughs in VLMs~\cite{chen2023pali,llava,beyer2024paligemma,wang2024qwen2} have spurred growing research interest in generalist action models.
These models~\cite{kim2024openvla,pi0,rdt,ghosh2024octo,wen2025diffusionvla,oxe,robovlm} use VLMs as the base model and are trained on diverse embodied intelligence datasets with the aim to achieve high levels of scenario, embodiment, and task generalization.
Methods such as SpatialVLA \cite{sptialVLA} enhance VLMs with 3D position embeddings for improved spatial reasoning. Despite these advances, VLM-based policies still face challenges in real-world performance and spatial generalization.

Our work focuses more on network design for improved 3D spatial understanding rather than concentrating on VLM-based generalist models. The most closely related approaches are KStarDiffuser~\cite{kstar} and SpatialVLA~\cite{sptialVLA}. Unlike SEM, both of them output EE poses rather than joint trajectories, and SpatialVLA does not model camera extrinsic or unify visual and action representations in a common 3D space.

\section{Problem Formulation}
The inputs for the manipulation problem primarily consist of sensor data and current joint positions $S$, where the sensor data generally encompass multi-view images $I$ and depth maps $D$. To simplify the problem, this paper utilizes only the sensor data from the current frame.
When handling multitasks, instruction texts $\mathrm{T}$ are also required.
Moreover, we consider modeling both the camera and the embodiment structure, thus requiring inputs of camera parameters $C$ and embodiment structure parameters $E$.
Thus, the inputs can be summarized into the following parts:
\begin{equation}
    \left\{ I \in \mathbb{R}^{N\times H\times W \times 3},
D \in \mathbb{R}^{N\times H\times W \times 1},
S\in \mathbb{R}^{N_{j}}, \mathrm{T} \right\} \cup \left\{C, E\right\}
\end{equation}
Here, $N$ denotes the number of images, $N_j$ represents the number of joints.
For a standard pinhole camera, $C$ represents the camera intrinsic and extrinsic matrix.
The output is the robot joint positions for the subsequence $t_{out}$ frames, denoted as $S_{out}\in \mathbb{R}^{N_{j}\times t_{out}}$.

\section{Our Method: SEM}

The overall architecture of SEM, illustrated in Fig.\ref{fig:overall_struture}, comprises five sub-modules: a vision language model (VLM) serving as a backbone for extracting features from images and text; a depth encoder for encoding 2D depth maps; a spatial enhancer that injects spatial and geometric awareness into 2D image tokens (Section~\ref{spatial_enhancer}); a robot state encoder used for encoding the robot's historical trajectory $S$ (Section~\ref{robot_state_encoder}); and an action decoder that predicts future trajectories $S_{out}$ from the aggregated features (Section~\ref{robot_action_decoder}).

\subsection{Backbone: VLM}
SEM can be integrated with any existing VLM to perform image and text feature encoding and leverage its strong semantic understanding capability. In this paper, we use two different types of VLMs: the 2D detection foundation model GroundingDINO and the LLM-based Qwen-2.5-VL.
Note that we consider GroundingDINO as a VLM in a broad sense.
The image features output by the VLM are denoted as $F^{I}$, and the text features are denoted as $F^{T}$.

\subsection{Spatial Enhancer}
\label{spatial_enhancer}

To effectively leverage both rich 2D image features and accurate 3D geometric information, we design a spatial enhancer within SEM. In multi-view feature fusion tasks such as 3D detection, a common practice (PETR~\cite{liu2022petr}, BIP3D~\cite{lin2025bip3d}) is to aggregate features based on sampled points along the camera frustum. Inspired by this, our spatial enhancer integrates multi-view 2D image features with depth-aware 3D representations to enhance 3D spatial understanding.

\begin{figure*}[tb]
     \centering
     \includegraphics[width=0.95\linewidth]{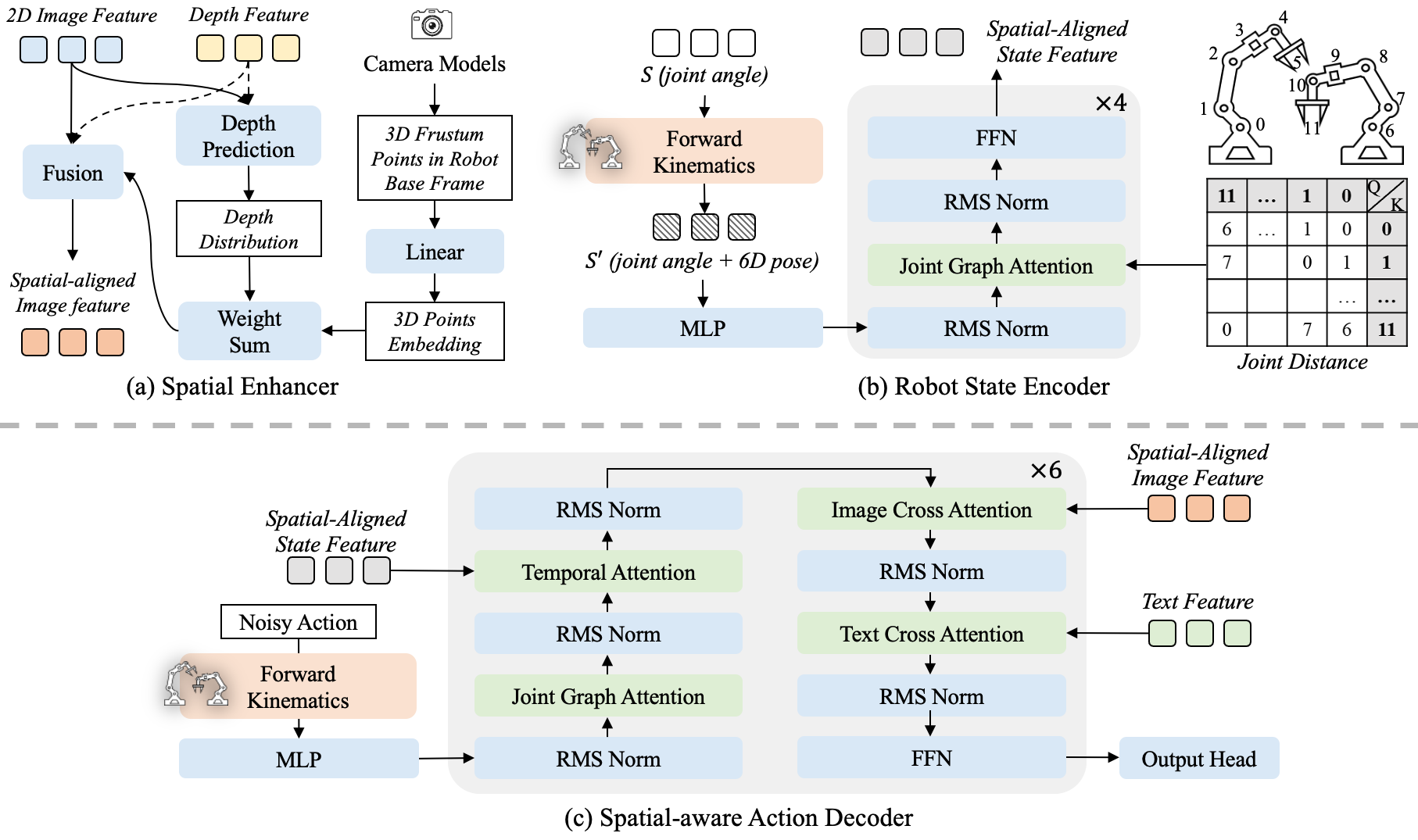}
     \vspace{-0.3cm}
     \caption{
     Details of SEM modules. (a) Spatial enhancer lifts multi-view features into the base-frame 3D space.
(b) Robot state encoder embeds joint angles and 6D poses with graph attention to model embodiment structure.
(c) Action decoder predicts future joint trajectories via diffusion.
     }
     \label{fig:robot_state_encoder}
 \end{figure*}

In this work, the proposed spatial enhancer performs two critical functions: (1) establishing a camera model to project 2D image features into 3D space, and (2) fusing image features with depth features to produce representations with stronger geometric awareness. As illustrated in Fig.~\ref{fig:robot_state_encoder}(a), the enhancer takes  multi-view image features, depth features, and camera parameters as input, and outputs enhanced features with better spatial understanding.
Specifically, for each image feature $F^{I}_{i,j}$, $K$ depth values is sampled, and the corresponding pixel coordinates are projected into 3D space using the camera model.
\begin{equation}
    P_{i,j} = \{T_{e}^{-1}T_{i}^{-1} [i \times d_k, j\times d_k, d_k]^{\bf{T}}  \in \mathbb{R}^{3} | 1\leq k \leq K\} 
\end{equation}
Here $T_{e}$ and $T_{i}$ are the extrinsic and intrinsic matrix, $i$ and $j$ are the pixel coordinates of image features. These 3D points $P$ are then encoded into high-dimension.
\begin{equation}
    E^{P}_{i,j,k} = {\bf MLP}(P_{i,j,k})
\end{equation}
Next, a discrete depth distribution $W$ is predicted and used to compute a weighted sum over these 3D points embeddings, resulting in a 3D positional embedding $E_{i,j}$ for each image feature.
\begin{equation}
    W_{i,j} = {\bf MLP}(F^I_{i,j}, F^D_{i,j})\in \mathbb{R}^{K} 
\end{equation}
\begin{equation}
    E_{i,j} = W_{i,j} \times E^{P}_{i,j}
\end{equation}
$F^D$ denotes the depth features output by the depth encoder. Finally, the image feature, depth feature, and 3D positional embedding are fused via a simple MLP.
\begin{equation}
    F^{3D}_{i,j} = {\bf MLP}(F^I_{i,j}, F^D_{i,j}, E_{i,j})
\end{equation}

To support scenarios without depth sensors, the depth feature is designed to be pluggable. When depth input is unavailable, we directly predict the depth distribution from image features, similar to monocular depth estimation.
For 3D encoding, we adopt the robot’s base coordinate frame to unify multi-view features. Subsequent 3D encodings for the embodiment are performed consistently in this coordinate frame, as detailed in Sections~\ref{robot_state_encoder} and~\ref{robot_action_decoder}.

\subsection{Robot State Encoder}
\label{robot_state_encoder}

The robot state encoder is used to encode the current robot state $S$.
A state representation containing only joint angles is insufficient, as the same joint angles may correspond to different physical states across embodiments.
To capture more informative information,
%
we compute the 6D pose of each joint in the robot’s base coordinate frame via forward kinematics and concatenate these poses with the joint positions to form the enhanced state representation $S'$.
\begin{equation}
    S'\in \mathbb{R}^{N_{j}\times 8}, S'_{i}=\left[a,x,y,z,q_w,q_x,q_y,q_z \right]_{i}
\end{equation}
where $a$ is the joint angle, $[x,y,z]$ represents the 3D coordinates of the joint, and $[q_w,q_x,q_y,q_z]$ denotes the rotation quaternion.
We project the feature $S'$ into a high-dimensional space using an MLP, then feed it into a transformer encoder with joint graph attention as its core component.
The joint graph attention utilizes the joint distance matrix $\bm{\mathrm{J}}\in \mathbb{R}^{N_j \times N_j\times 1}$ to encode relative positions, where each entry $\bm{\mathrm{J}}_{i,j}$ denotes the number of links between the $i_{th}$ query joint and the $j_{th}$ key joint.
The formulation is as follows.
\begin{equation}
    \bm{\mathrm{P}} = \mathrm{MLP}(\mathrm{SinCos} (\bm{\mathrm{J}}) ) \in \mathbb{R}^{N_j \times N_j \times d}
\end{equation}
\begin{equation}
    \bm{\mathrm{A}} = \mathrm{einsum}(ik,ijk,jk\rightarrow ij | \bm{\mathrm{Q}}, \bm{\mathrm{P}}, \bm{\mathrm{K}}) \in \mathbb{R}^{N_j \times N_j}
\end{equation}
\begin{equation}
    \bm{\mathrm{O}} = \mathrm{softmax}\left(\frac{\bm{\mathrm{A}}}{\sqrt{d}}\right) \bm{\mathrm{V}}
\end{equation}
where $\bm{\mathrm{Q}}$, $\bm{\mathrm{K}}$, $\bm{\mathrm{V}}$ and $\bm{\mathrm{O}}$ denote query, key, value and output features, $d$ is the feature channels, and $\mathrm{einsum}$ is the Einstein summation convention.
It is important to note that the computational complexity of joint graph attention is $O(N_j^2 d^2)$, which is higher than vanilla attention's $O(N_j^2 d+N_j d^2)$. However, since $N_j$ is typically small, the computational load of this module is not a bottleneck.

It can be observed that our robot state encoder is joint-centric and naturally generalizes to embodiments with an arbitrary number of joints, which is crucial for training with mixed multi-embodiment data.

\subsection{Spatial-Aware Action Decoder}
\label{robot_action_decoder}
The action decoder takes all condition features (image features, text features, and robot state features) as input and employs a diffusion transformer to predict future trajectories.

First, we sample noise on the joint angles. Then, similar to the robot state encoder, we use forward kinematics to obtain the noisy robot state, denoted as $S_{noise}' \in \mathbb{R}^{N_j \times t_{out} \times 8}$.
In terms of network architecture, the action decoder shares a joint-centric design with the state encoder, utilizing joint graph attention to enable feature interaction among joints. 
We employ three independent cross-attention layers that attend to robot state, image, and text features respectively (see Fig.~\ref{fig:robot_state_encoder}(c)). The temporal cross-attention operates in a causal manner, ensuring a unidirectional flow along the time axis. Finally, an output head consisting of upsampling and 1D convolution layers generates the structured prediction $S_{pred}' \in \mathbb{R}^{N_j \times t_{out} \times 8}$, which can be further refined via denoising steps.
For motion control, we use only the predicted joint angles in $S_{pred}'$, while the predicted 6D poses just serve as auxiliary supervision. Nevertheless, future work could explore leveraging the predicted 6D poses with inverse kinematics for control.


\begin{figure*}[tb]
     \centering
     \includegraphics[width=1\linewidth]{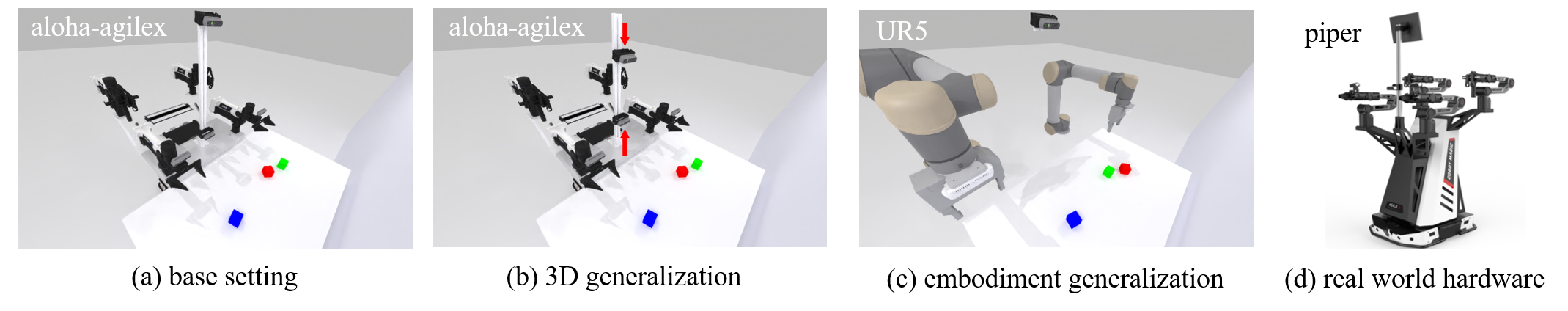}
     \vspace{-0.8cm}
     \caption{We use RoboTwin2.0 as the simulation benchmark. The base environment features the AgileX-Aloha embodiment, equipped with two environment cameras and two hand cameras, on a clean table. For the 3D generalization experiments, we adjust the height of the two environment cameras. For the embodiment generalization experiments, we replace the test robot with two UR5 arms. In real-world testing, we use the Piper.}
     \label{fig:eval_setting}
 \end{figure*}

 \begin{table*}[h]
    \caption{Model parameter counts (M). GD denotes GroundingDINO Tiny, QW represents Qwen2.5-VL-3B. Parameters marked with underscores remain frozen during training. \scalebox{1.5}{$\epsilon$}, $E$, and $D$ denote the encoders, enhancers, and decoder respectively.}
    \centering
    \begin{tabular}{l|ccccc|cc}
    \toprule
    Model & VLM & \scalebox{1.5}{$\epsilon$}$_{depth}$  &  \scalebox{1.5}{$\epsilon$}$_{state}$ & $E_{spatial}$ & $D_{action}$ & Trainable & All \\
    \midrule
    SEM - GD & 153.00  & 0.09  & 5.77 & 0.73  & 15.55 &  175.33 &  175.33\\
    SEM - QW & \underline{3580.69} & 0.74 & 5.77 & 1.25 &15.55& 41.33 & 3622.02\\
    \bottomrule
    \end{tabular}
    \label{tab:param}
\end{table*}

\subsection{Training Loss}
\label{robot_loss}
During training, we optimize four loss terms: joint position loss, joint pose loss, forward kinematics pose loss, and depth loss, as defined in Equation~\ref{loss}.

\begin{equation}
\label{loss}
    L = \lambda_1 L_{joint} + \lambda_2 L_{pose} + \lambda_3 L_{pose}^{fk} + \lambda_4 L_{depth}
\end{equation}
Here, the joint position loss $L_{joint}$ and joint pose loss $L_{pose}$ are defined as the L1 distances between the predicted joint angles and 6D poses and their respective ground truths. To further improve positional accuracy, the forward kinematics pose loss $L_{pose}^{fk}$  recomputes 6D poses from the predicted joint angles using forward kinematics and compares them with the ground truth via an additional L1 distance.
Unlike prior methods, our approach simultaneously constrains joint angles and 6D poses, providing stronger and more informative supervision, which  leads to improved performance. The depth loss $L_{depth}$, applied to the depth distribution predicted by the spatial enhancer, is formulated as a cross-entropy loss.

\section{Experiments}
\subsection{Benchmark}

We evaluate SEM in both simulation and real-world settings. In simulation, we conduct experiments on the RoboTwin 2.0 platform~\cite{robotwin2} using the aloha-agilex embodiment.  
From RoboTwin’s task suite, we select 16 representative tasks based on diversity, difficulty, and expert trajectory synthesis success rate. The selected tasks are listed in the Tab.\ref{tab:main_ret}.
To assess SEM’s generalization capabilities, we design two additional evaluations: (1) a 3D spatial generalization test and (2) an embodiment generalization test.  
Both are conducted in simulation, as illustrated in Fig.~\ref{fig:eval_setting}(b,c), with details provided in Sec.~\ref{3d generalization} and Sec.~\ref{embodiment generalization}.  
Real-world experiments are presented separately in Sec.~\ref{sec:realworld}. 

\subsection{Implementation Details}

We adopt Grounding DINO Tiny~\cite{liu2024grounding} and Qwen2.5-VL-3B~\cite{qwen2.5-vl} as the VLM backbone. When using Qwen2.5-VL-3B, we freeze its parameters and train only the newly added modules. The parameter distributions of both configurations are shown in Tab.\ref{tab:param}.
Compared with most existing VLA models~\cite{bjorck2025gr00t,pi0,rdt}, SEM uses a lightweight action decoder with substantially fewer parameters. 
%
%
This aligns with a hierarchical design philosophy: the VLM backbone acts as the "brain," providing semantic understanding and reasoning, while the action decoder is responsible for instruction following, and thus requires comparatively fewer parameters.

For training, we use 160k steps for multi-task models and 80k steps for single-task models, with 256 batch size.
For the diffusion process, we adopt DDPM~\cite{ho2020denoising} with 1000 timesteps and use DPM-Solver~\cite{lu2022dpm} for fast inference with 10 denoising steps.  
We directly predict the noisy sample rather than noise, which empirically accelerates convergence.



\subsection{Main Results}
\label{main results}
We first evaluate our method under the single-task setting, with each model trained on 50 demonstrations per task.
As shown in Tab.~\ref{tab:main_ret}, SEM achieves significant performance gains, improving average success rates by 22.13\% over DP3~\cite{dp3}, 35.38\% over ACT~\cite{aloha} and 47.32\% over DP~\cite{dp}.
Compared to two well-pretrained models, RDT~\cite{rdt} and Pi0~\cite{pi0}, SEM also shows significant advantages in the simulation environment.
Notably, SEM demonstrates especially strong performance on tasks requiring high-precision manipulation, achieving an 80.0\% success rate on the `block ranking rgb' task, compared to 3.0\% (DP3) and 0.0\% (DP).

We further assess performance in multi-task scenarios, training each model on 1,600 demonstrations (100 per task).
We evaluated two variants of our model, SEM-GD and SEM-QW, achieving success rates of 84.5\% and 84.62\%, respectively, both significantly outperforming the baseline RDT-1B (55.12\%). Meanwhile, our trainable parameters are also considerably fewer than RDT-1B.

\input{main_results_tab}

\begin{figure*}[htbp]
  \centering
  \begin{minipage}{0.635\textwidth}
    \includegraphics[width=\linewidth]{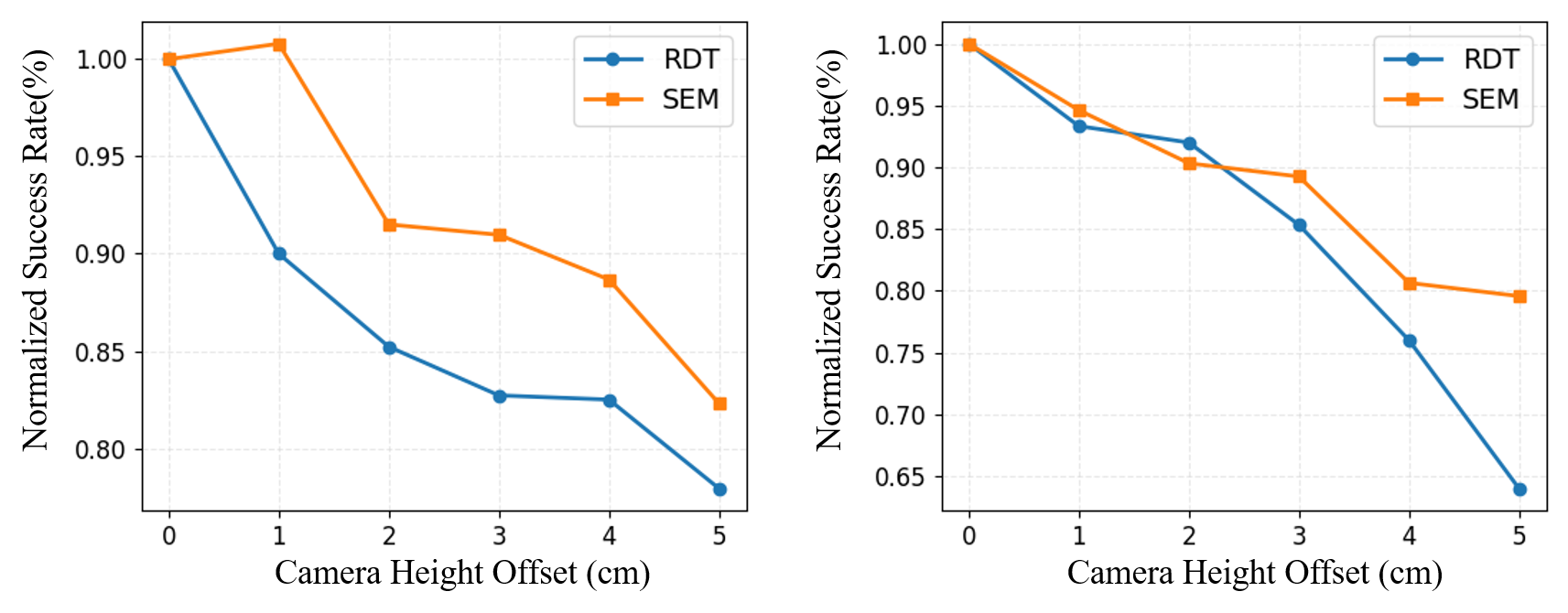}
    \caption{3D Generalization Test Results. The normalized success rate is defined as the ratio of at each camera height offset to at 0 cm. Left: Multi-task \& multi-view. Right: Single-view, single-task (place empty cup).}
    \label{fig:3d_gen_ret}
  \end{minipage}
  \hfill
  \begin{minipage}{0.31\textwidth}
    \hspace*{-0.3cm}%
    \includegraphics[width=\linewidth]{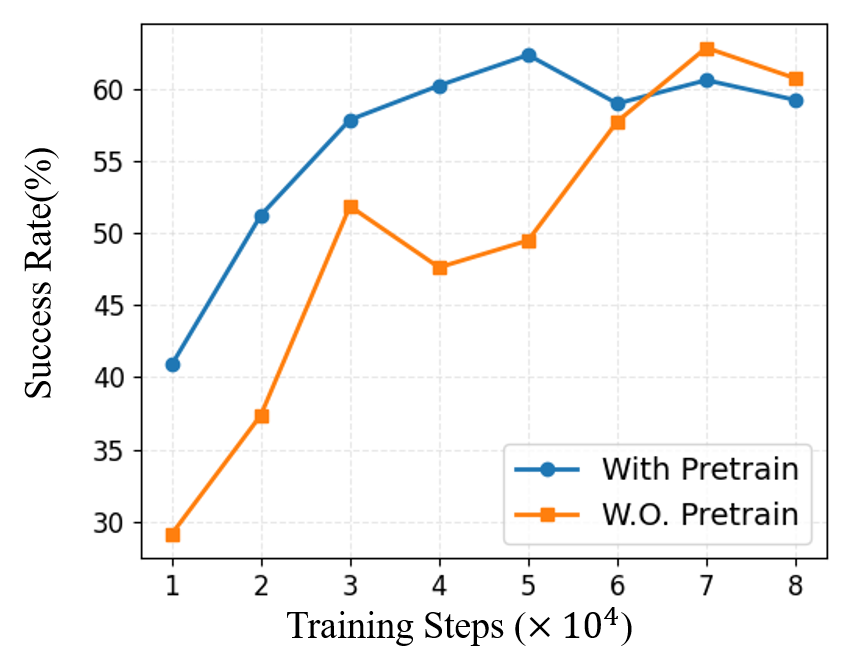}
    \caption{Embodiment Generalization Results. Performance on UR5, with and without pretraining on Agilex.}
    \label{fig:emb_gen}
  \end{minipage}
\end{figure*}

\subsection{3D Generalization Test}
\label{3d generalization}
To demonstrate that SEM possesses strong 3D generalization capabilities, we design the following experiment. We adjusted the height of the two environment cameras, as shown in Fig.\ref{fig:eval_setting}(b), and then conduct zero-shot testing of the multi-task model SEM-QW and RDT under this modified camera setup. SEM consistently gets higher average success rates across 16 tasks compared to RDT, and its normalized success rate exhibits a slower rate of decay as camera height varies, as shown in Fig.\ref{fig:3d_gen_ret}(left).

Since some tasks  (e.g., handover mic and open laptop) are not sensitive to camera height variations, and the presence of hand cameras may influence the experimental conclusions, we also train two single-task (place empty cup) models with a single environment camera, and test them under different camera heights. SEM also demonstrate superior 3D generalization capabilities in this setting, as shown in Fig.\ref{fig:3d_gen_ret}(right).

\input{ablation_tab}
\begin{figure*}[tb]
     \centering
     \includegraphics[width=1\linewidth]{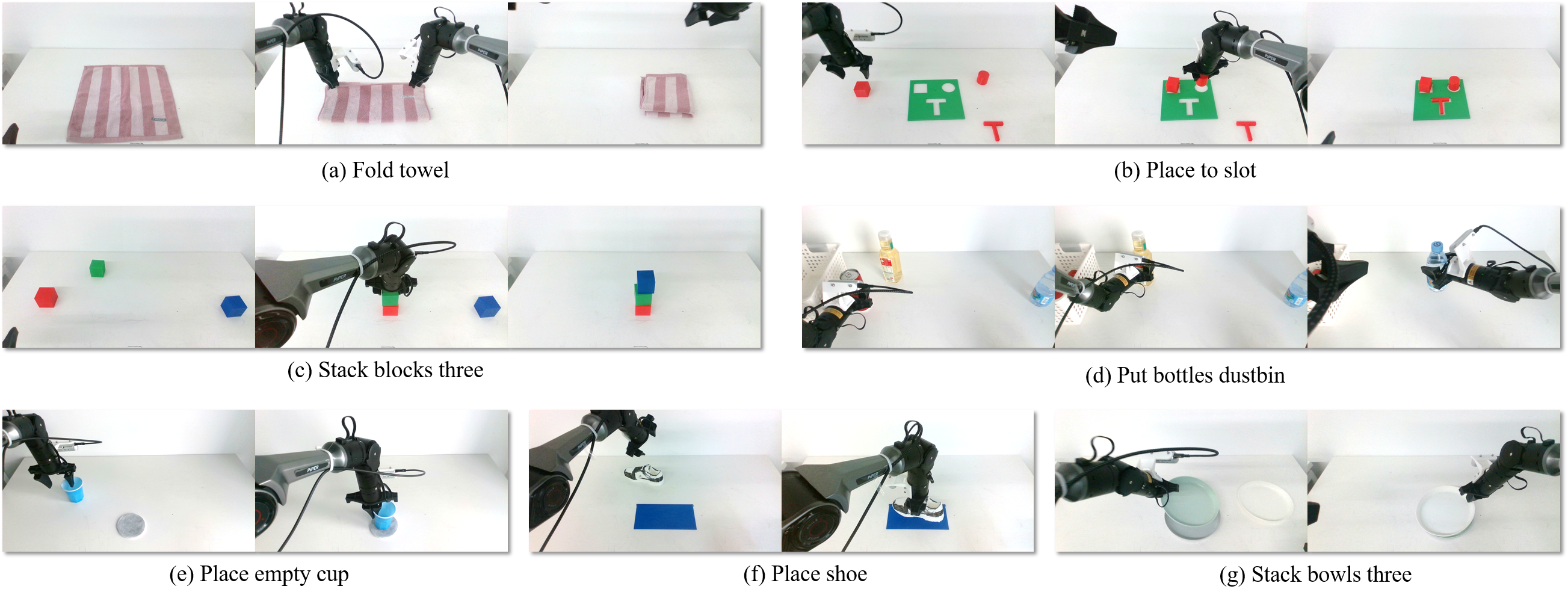}
     
     \vspace{-0.4cm}
     \caption{
     Examples of real-world tasks. (a) deformable object manipulation task, (b) fine manipulation task, (c–g) same as RoboTwin settings.
     }
     \label{fig:realword_pic}
 \end{figure*}

 \begin{table}[htbp]
  \centering
  
  \caption{Embodiment Generalization Results. Training with multi-embodiment data yields performance gains for SEM, but has a significant negative effect on RDT.}
  \vspace{-0.2cm}
  \begin{tabular}{c|c|c}
  
    \toprule
     & SEM & RDT\\
     \midrule
        Only UR5 & 60.75 & 63.00\\
        Agilex pretrain, UR5 finetune & 59.25 (\bf{-1.50}) & 46.30 (\bf{-16.70})\\
        Mix Agilex and UR5 & 66.25 (\bf{+5.50}) & 36.31 (\bf{-26.69})\\
        \bottomrule
    \end{tabular}
    \label{tab:emb_gen}
\end{table}

 \begin{table}[htbp]

  \centering
  \caption{Task progress in the real-world testing. The ``+" indicates models pretrained on open-source real-world data, while the others were pretrained on RoboTwin simulation data.}
    \begin{tabular}{l|c|c|c|c}
        \toprule
         & Task a & Task b & Task c &   Task d \\
         \midrule
         RDT & - & - & 44.0 &  36.0\\
         \rowcolor{gray!20}  SEM & - & - & 66.0 & 38.0 \\
         RDT$^+$ & 41.3 & 68.0 & 64.0 & 33.0 \\
         \rowcolor{gray!20}  SEM$^+$  & 68.4 & 69.0 & 78.0 & 68.0 \\
        \midrule
        & Task e & Task f & Task g & Mean\\
        \midrule
        RDT & 42.7 & 18.0 & 65.3 & 41.2 \\
         \rowcolor{gray!20} SEM & 63.0 & 34.0 & 66.7 & 53.5 (\bf{+12.3})\\
        RDT$^+$  & 38.0 & 28.0 & 55.0 & 46.8\\
         \rowcolor{gray!20} SEM$^+$  & 64.7 & 64.7 & 95.3 & 72.6 (\bf{+25.8})\\
        \bottomrule
    \end{tabular}
    \label{tab:real world}
\end{table}


\subsection{Embodiment Generalization Test}
\label{embodiment generalization}
To evaluate the embodiment generalization capability, we train on a mixed dataset consisting of data from two robot embodiments: AgileX-Aloha and Dual-UR5. Specifically, it includes 100 demonstrations per task for Agilex-Aloha and 50 demonstrations per task for UR5. As a comparison, we train the model using only UR5 data, with 50 demonstrations per task. If the model possesses embodiment generalization ability, the inclusion of AgileX-Aloha data is expected to improve its performance on the UR5.
The results, shown in Tab.\ref{tab:emb_gen}, indicate that training with the mixed dataset leads to a 5.5\% performance improvement for SEM compared to training with UR5 data only, whereas RDT exhibits a performance drop of 26.69\%.
SEM possesses strong embodiment generalization capability, whereas models like RDT can be negatively affected by the ambiguity of multi-embodiment data.
In addition, we finetune Agilex-pretrained SEM on the UR5 dataset and observe a significant improvement in convergence speed, as shown in Fig.\ref{fig:emb_gen}.


\subsection{Ablation Study}
\label{ablation study}
We first conduct an ablation study on the spatial enhancer. This module serves two main functions: fusing depth features and performing 3D encoding. Removing only the depth features reduced the average success rate by 8.5\%, and further removing the entire spatial enhancer led to a total reduction of 15.72\%, as shown in Tab.\ref{tab:ablation} (id 1 \& 2).

We also ablate the unified 3D space. Replacing the image 3D position embedding from the unified base frame with the camera frame reduces performance by 3.37\% (Tab.\ref{tab:ablation}, id 3).

Last, we evaluated the role of the joint graph attention by replacing it with vanilla attention, which resulted in a 4.62\% reduction (Tab.\ref{tab:ablation}, id 4).
It is important to highlight that joint graph attention is expected to be even more pronounced when trained on more diverse embodiment data, which we aim to verify in future work.

\subsection{Real World Experiments}
\label{sec:realworld}
%
%
%

For the real-world experiments, we evaluate SEM on seven tasks.  
Five tasks are selected from RoboTwin: as shown in Fig.~\ref{fig:realword_pic} (c-g). %
In addition, we introduce two novel tasks: \textit{place blocks into slots}, designed to evaluate fine-grained manipulation capability, and \textit{fold towel}, which evaluates performance on deformable object manipulation.  
The experimental setup is shown in Fig.~\ref{fig:realword_pic}.  
We conduct two sets of comparative experiments:  
(1) both SEM and RDT are first pretrained on RoboTwin simulation data and then post-trained on real-world demonstrations for a fair comparison;  
(2) following common practice, SEM is pretrained on the open-source real-world dataset (Agibot-world we used) and post-trained on our data, while RDT uses its open-source pretrained weights. 
As summarized in Tab.\ref{tab:real world}, SEM significantly outperforms RDT whether pretrained on simulation or real-world data.

\section{Conclusion and Future Works}

This work introduces SEM, a vision-language-action framework that builds a unified spatial representation by projecting multi-view image features and robot states into a shared 3D base frame.
By explicitly modeling camera parameters and embodiment structure, SEM enables robust, generalizable end-to-end motion planning.
%
Extensive experiments in both simulation and real-world settings, along with ablation studies, validate the effectiveness of SEM in improving 3D spatial and embodiment generalization.


Despite these results, SEM still has room for further improvement. Future work includes refining the definition of relative joint distances, integrating more robust and accurate depth prediction model, pretraining on larger and more diverse datasets, leveraging VLMs for reasoning over long-horizon tasks, and exploring reinforcement learning to improve policy robustness and reliability.




\bibliographystyle{IEEEtran}
\bibliography{root}

\end{document}

%% file: main_results_tab.tex
\begin{table*}[tb]
    \centering
    \caption{RoboTwin2.0 Benchmark Results. * indicates multi-task models, while others are single-task models. Single-task results of baselines are all from the official RoboTwin leaderboard.}
    \vspace{-0.2cm}
    \begin{tabular}{l|c|c|c|c|c|c}
        \toprule
         & adjust bottle & beat block hammer & blocks ranking rgb & blocks ranking size & dump bin bigbin & handover mic  \\
         \midrule
        DP~\cite{dp} & 97.00 & 42.00 & 0.00 & 1.00 & 49.00 & 53.00  \\
        RDT~\cite{rdt} & 81.00 & 77.00 & 3.00 & 0.00 & 64.00 & 90.00 \\
        ACT~\cite{aloha} & 97.00 & 56.00 & 1.00 & 0.00 & 68.00 & 85.00  \\
        Pi0~\cite{pi0} & 90.00 & 43.00 & 19.00 & 7.00 & 83.00 & 98.00   \\
        DP3~\cite{dp3} & 99.00 & 72.00 & 3.00 & 2.00 & 85.00 & 100.00     \\
        
        \rowcolor{gray!20}  SEM-QW & 100.00 & 86.00 & 80.00 & 12.00 & 96.00 & 100.00      \\
        RDT*  & 98.00 & 84.00 & 20.00 & 3.00 & 92.00 & 100.00    \\
        \rowcolor{gray!20} SEM-GD*  & 98.00 & 100.00 & 66.00 & 40.00 & 100.00 & 100.00  \\
        \rowcolor{gray!20}  SEM-QW*  &100.00 & 96.00 & 82.00 & 34.00 & 98.00 & 100.00   \\
        \midrule
        
        & lift pot & move pillbottle pad & open laptop & open microwave & place cans plasticbox & place dual shoes \\
        \midrule
       DP  & 39.00 & 1.00 & 49.00 & 5.00 & 40.00 & 8.00\\ 
       RDT & 72.00 & 8.00 & 59.00 & 37.00 & 6.00 & 4.00 \\
       ACT & 88.00 & 0.00 & 56.00 & 86.00 & 16.00 & 9.00 \\
       Pi0 & 84.00 & 21.00 & 85.00 & 80.00 & 34.00 & 15.00 \\
       DP3  & 97.00 & 41.00 & 82.00 & 61.00 & 48.00 & 13.00 \\ 
       
       \rowcolor{gray!20} SEM-QW &  92.00 & 62.00 & 84.00 & 96.00 & 84.00 & 48.00\\
        RDT* & 80.00 & 13.00 & 61.00 & 59.00 & 16.00 & 14.00\\
       \rowcolor{gray!20} SEM-GD*  & 98.00 & 68.00 & 82.00 & 82.00 & 100.00 & 78.00\\
       \rowcolor{gray!20}  SEM-QW* & 96.00 & 70.00 & 86.00 & 98.00 & 94.00 & 58.00 \\
       \midrule
        & place empty cup & rotate qrcode & stack blocks three & stack bowls three &Mean& Gain \\
        \midrule
        DP & 37.00 & 13.00 & 0.00 & 63.00 & 31.06 \\
        RDT & 56.00 & 50.00 & 2.00 & 51.00 & 41.25\\
        ACT & 61.00 & 1.00 & 0.00 & 48.00 & 42.00\\
        Pi0 & 37.00 & 68.00 & 17.00 & 66.00 & 52.94 \\
        DP3 & 65.00 & 74.00 & 1.00 & 57.00 &56.25\\
        
        \rowcolor{gray!20} SEM-QW & 96.00 & 84.00 & 54.00 & 80.00 &78.38 & \bf{+22.13}\\
         RDT* &  79.00 & 82.00 & 17.00 & 64.00 & 55.12 \\
         \rowcolor{gray!20} SEM-GD*  & 94.00 & 86.00 & 80.00 & 80.00 &84.50 &\bf{+29.38} \\
        \rowcolor{gray!20}  SEM-QW* &98.00 & 88.00 & 72.00 & 84.00 & 84.62&\bf{+29.50}\\
        \bottomrule
    \end{tabular}
    
    \label{tab:main_ret}
\end{table*}

%% file: ablation_tab.tex
\begin{table*}[tb]
\centering
\caption{Ablation study of SEM. 'SE' and 'JGA' denote 'spatial enhancer' and 'joint graph attention', respectively.}
\vspace{-0.2cm}
\label{tab:ablation}
    \begin{tabular}{l|l|c|c|c|c|c|c}
        \toprule
         id & setting& adjust bottle & beat block hammer & blocks ranking rgb & blocks ranking size & dump bin bigbin & handover mic  \\
         \midrule
         1&No Depth \& SE &100.00 & 90.00 & 44.00 & 4.00 & 90.00 & 100.00  \\
         2&No Depth & 100.00 & 96.00 & 76.00 & 24.00 & 94.00 & 100.00   \\
         3 & No Unified 3D & 100.00 & 88.00 & 84.00 & 28.00 & 92.00 & 100.00 \\
         4 & No JGA & 100.00 & 98.00 & 78.00 & 46.00 & 92.00 & 100.00  \\
         \rowcolor{gray!20} &  Full Model &100.00 & 96.00 & 82.00 & 34.00 & 98.00 & 100.00 \\
        \midrule
        id & setting & lift pot & move pillb. pad & open laptop & open micro. & place cans plast. & place shoes \\
        \midrule
       1&No Depth \& SE & 82.00 & 20.00 & 70.00 & 56.00 & 44.00 & 32.00 \\
       2&No Depth & 84.00 & 48.00 & 80.00 & 86.00 & 80.00 & 44.00 \\
       3 & No Unified 3D & 96.00 & 50.00 & 88.00 & 96.00 & 100.00 & 58.00\\
         4&No JGA & 94.00 & 58.00 & 80.00 & 66.00 & 100.00 & 58.00 \\
         \rowcolor{gray!20}  &Full Model & 96.00 & 70.00 & 86.00 & 98.00 & 94.00 & 58.00\\
       \midrule
        id & setting & place empty cup & rotate qrcode & stack blocks three & stack bowls three & Mean & Degradation \\
        \midrule
        1&No Depth \& SE &  72.00 & 80.00 & 34.00 & 64.00 & 58.80 & \bf{-15.72}\\
        2&No Depth & 94.00 & 80.00 & 56.00 & 76.00 & 76.12 & \bf{-8.50} \\
         3 & No Unified 3D &  94.00 & 92.00 & 50.00 & 84.00 & 81.25 & \bf{-3.37} \\
        4& No JGA & 98.00 & 86.00 & 50.00 & 76.00 & 80.00 & \bf{-4.62}\\
        \rowcolor{gray!20} &  Full Model & 98.00 & 88.00 & 72.00 & 84.00 & 84.62 \\
        \bottomrule
    \end{tabular}
    
\end{table*}